\documentclass[twocolumn, switch]{article} 

\usepackage{preprint}

\usepackage{amsmath, amsthm, amssymb, amsfonts}
\usepackage{subcaption}
\usepackage[numbers,square]{natbib}
\usepackage[utf8]{inputenc}	
\usepackage[T1]{fontenc}	
\usepackage{xcolor}		
\usepackage[colorlinks = true,
            linkcolor = purple,
            urlcolor  = blue,
            citecolor = cyan,
            anchorcolor = black]{hyperref}	
\usepackage{booktabs} 		
\usepackage{nicefrac}		
\usepackage{microtype}		
\usepackage{lineno}		
\usepackage{float}			

\usepackage{newfloat}
\DeclareFloatingEnvironment[name={Supplementary Figure}]{suppfigure}
\usepackage{sidecap}
\sidecaptionvpos{figure}{c}

\usepackage{titlesec}
\titlespacing\section{0pt}{12pt plus 3pt minus 3pt}{1pt plus 1pt minus 1pt}
\titlespacing\subsection{0pt}{10pt plus 3pt minus 3pt}{1pt plus 1pt minus 1pt}
\titlespacing\subsubsection{0pt}{8pt plus 3pt minus 3pt}{1pt plus 1pt minus 1pt}

\usepackage{tikz,xcolor,hyperref}

\definecolor{lime}{HTML}{A6CE39}
\DeclareRobustCommand{\orcidicon}{
	\begin{tikzpicture}
	\draw[lime, fill=lime] (0,0) 
	circle [radius=0.16] 
	node[white] {{\fontfamily{qag}\selectfont \tiny ID}};
	\draw[white, fill=white] (-0.0625,0.095) 
	circle [radius=0.007];
	\end{tikzpicture}
	\hspace{-2mm}
}
\foreach \x in {A, ..., Z}{\expandafter\xdef\csname orcid\x\endcsname{\noexpand\href{https://orcid.org/\csname orcidauthor\x\endcsname}
			{\noexpand\orcidicon}}
}

\title{Comparative Evaluation of Applicability
Domain Definition Methods for Regression Models}

\usepackage{authblk}

\author[1]{Shakir Khurshid}
\author[2]{Bharath Kumar Loganathan}
\author[3]{Matthieu Duvinage}

\affil[1]{Department of Computer Science, Sapienza University of Rome}
\affil[2]{Cognizant Technology Solutions, India}
\affil[3]{GSK , Wavre}

\begin{document}
\twocolumn[ 
  \begin{@twocolumnfalse} 
  
\maketitle

\begin{abstract}
The applicability domain refers to the range of data for which the prediction
of the predictive model is expected to be reliable and accurate and using a model
outside its applicability domain can lead to incorrect results. The ability to define the regions in data space where a predictive model can be safely used is a necessary condition for having safer and more reliable predictions to assure the reliability of new predictions. However, defining the applicability domain of a model is a challenging problem, as there is no clear and universal definition or metric for it. This work aims to make the applicability domain more quantifiable and pragmatic. Eight appicability domain detection techniques 
were applied to seven regression models, trained on five different datasets, and their performance was benchmarked using a validation framework. We also propose a novel approach based on non-deterministic Bayesian neural networks to define the applicability domain of the model. Our method exhibited superior accuracy in defining the Applicability Domain compared to previous methods, highlighting its potential in this regard.

\end{abstract}
\vspace{0.35cm}

  \end{@twocolumnfalse} 
] 



\section{Introduction}
Software systems utilizing predictive modelling are used in multiple industries.
However, when deployed in a production environment there is no definitive way
to measure the reliability of their predictions. So the question arises "Can their prediction be trusted ?". Currently, the main way of evaluating a model is by measuring its test error however in the real-time environment we do not have a test set to measure the accuracy of the predictions, thus we need alternative methods to validate the predictions.

The primary motivation to undergo this work was to get a quantifiable metric that can measure the reliability of the predictions of the prediction model used in the vaccine production process. At GSK in the vaccine manufacturing process, sensors installed in fermenters used in the vaccine manufacturing process collect data such as dissolved oxygen, pressure level and temperature. This data is then fed into a machine-learning model that predicts the yield of the vaccine for that batch.
However, these models are typically developed based on historical data, process parameters, and specific conditions under which the vaccines were produced. However, the production environment is subject to variability, including changes in raw materials, equipment performance, and environmental conditions, which may affect the accuracy of the predictions, causing a risk of overestimating or underestimating the production capacity, leading to vaccine shortages, inefficient resource allocation or supply chain disruptions Without a clear understanding of the limits and boundaries of our models, there is a risk of blindly applying them to scenarios for which they may not be suitable or reliable. This lack of awareness can lead to erroneous predictions, faulty decision-making, and potential negative consequences. Thus, defining a model’s applicability domain allows us to gain insight into the model’s scope and limitations, enabling us to make safer predictions in real time.

The main questions that this research seeks to address are:
1. How can we effectively define and determine the applicability domain of a model?
2. Establish a robust evaluation framework to compare and assess the performance of different techniques to define the applicability domain.
To answer these questions, we will investigate several existing methods for detecting and quantifying AD in prediction models, including statistical metrics, distribution-based approaches, and deep learning-based techniques. We will also propose a new method based on a Bayesian neural network to do the same.




\section{Applicability Domain}

The applicability domain of a model refers to the region or range of input data where the model’s predictions are expected to be reliable and accurate. It represents the domain or space of inputs for which the model is deemed appropriate or applicable. The concept of the applicability domain is important because models are typically trained on specific types or ranges of data, and their performance may vary outside of that range. The applicability domain helps us understand the limitations and validity of a model’s predictions based on the input data characteristics. It can be influenced by various factors, including the nature of the data used for training the model, the complexity of the model, and the assumptions made during model development. It can be defined in different ways, depending on the specific context and requirements of the model. If a new data point falls within the applicability domain, it is expected that the model’s predictions will be reliable for that point. However, if a data point falls outside the applicability domain, the model’s predictions may be less accurate or less reliable. Different techniques can be employed to
assess the applicability domain, such as measuring the similarity of new data to the training data using distance metrics, examining the distributional characteristics of the data, or using domain-specific knowledge and expert judgment. Understanding the applicability domain of a model is crucial for making informed decisions about its usage, identifying potential limitations or areas of high uncertainty, and ensuring that the model is applied in appropriate contexts where its predictions are reliable and meaningful
\begin{figure}[th]
\centering
\includegraphics[width=8cm]{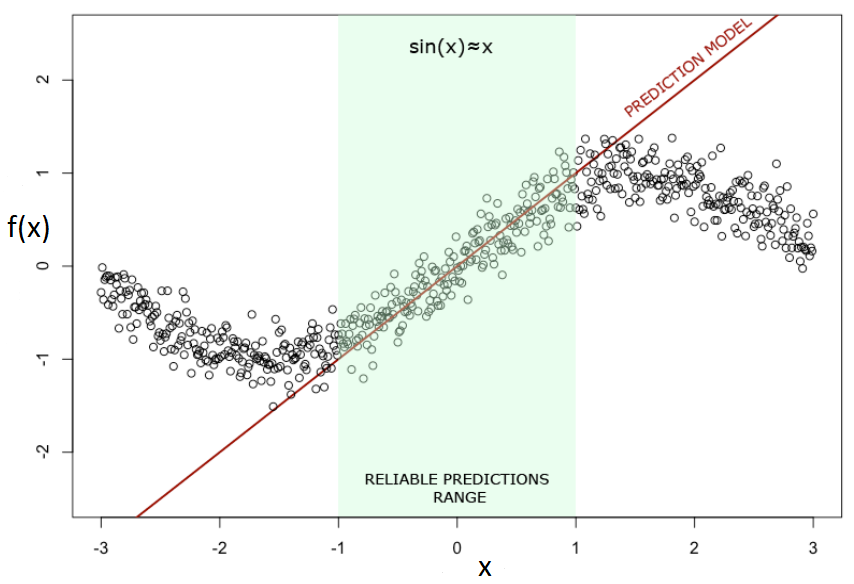}
\caption[Illustrative example of applicability domain]{ An illustrative example \cite{Reference1} for the applicability domain problem. Within the green region, a linear model (represented by the red line) provides a good approximation of the data. However, outside the green region, the linear model's approximation is not valid. As a result, the applicability domain of the linear model is defined by the interval [-1, 1], which corresponds to the green region.}
\label{fig:AD}
\end{figure}

\subsubsection{Applicability Domain Measures}\label{sec:AD measures}
Sushko et al. \cite{Reference1}  introduced the term Distance to Models, which is an abstract concept that is used to define the Applicability domain. It is not an actual distance. It represents a metric measure that defines the similarity between the training set and test set data points or unseen inputs (in the production phase) for a given predictive model. It is defined as monotonically increasing as the (expected) accuracy of the model decreases. However, this term is misleading since most measures are not distances. Therefore to avoid potential confusion we used the term \textit{Applicability Domain Measure} \cite{Reference2} instead of Distance to Model. In accordance with Sushko et al. \cite{Reference1}, the data points which have larger values of AD measure are by definition expected to have lower prediction accuracy than data points that have smaller values of AD measure. It should be clearly stated that prediction accuracy correlates with AD measure only on average: for example, data points in the range [0.1, 0.3] generally exhibit higher prediction accuracy than those in the range [0.6, 0.7]. However, individual data points within the first interval may still have larger prediction errors than some data points within the second interval. In other words, the key property of an applicability domain (AD) measure is its discriminating ability, distinguishing between high and low-accuracy predictions. Predictions falling below a predefined AD threshold are considered within the domain. The threshold can be manually chosen, and efficient AD measures exhibit a monotonically increasing relationship between error rate and AD value. These measures estimate the reliability of predictions, a subjective aspect not as objectively defined as accuracy. 
Various AD measures exist, evaluating prediction reliability from different perspectives. Our focus includes two categories: \textit{Novelty detection}, relying solely on input data, and \textit{Confidence estimation}, utilizing information from the underlying model, which proves more potent for AD definition. 

\subsubsection{Novelity Detection}
\paragraph{DA-Index}
 This method is based on the K-NN approach. We used K=5 and Euclidean distance was used as a distance measure. A lower DA\_index value indicates greater similarity between test and train data. DA index itself comprises four measures. $\kappa, \gamma, \delta$ \cite{Reference2}  and we added another measure to it named min-$\kappa$. $\kappa$ represents the distance of the test point to the kth-nearest neighbour in the training set. min-$\kappa$ represents the minimum distance from the test point to the k-nearest neighbours. $\gamma$ represents the mean distance of a test point to its k-nearest neighbours and $\delta$ corresponds to the length of the mean vector from a test point to its k-nearest neighbours. 

\paragraph{Cosine Similarity}
 This measure \cite{Reference2} measures the cosine similarity between the test point and its k-nearest training set neighbours. In this study, we chose k to be 5. Cosine similarity between two data points $x_a$ and $x_b$ is determined by taking the inner product of the two vectors and dividing it by the product of their vector lengths.
\begin{equation}
\cos(\alpha_{x_a, x_b}) = \frac{\sum_{i=1}^{p} x_{a, i} \cdot x_{b, i}}{\sqrt{\sum_{i=1}^{p} x_{a, i}^2 \sum_{i=1}^{p} x_{b, i}^2}}
\end{equation}
This represents the angle between two vectors originating from the origin and extending to the two $x_a$ and $x_b$  p-dimensional data points.
The cosine value ranges between 0 and 1, with a value of 1 indicating perfect similarity. The expression $1 -  \cos(\alpha_{x_a, x_b})$ was used to convert the cosine from a similarity measure to an AD Measure (i.e. a dissimilarity measure)  

\paragraph{Leverages}
This method employs the Mahalanobis distance to measure the proximity of a data point to the centre of the training set distribution. The leverage, denoted as h, is computed using the "hat" matrix through the equation $h = (x_i^T(X^TX)^{-1})x_i)$, where X represents the matrix of the training set data points and $x_i$ represents a specific data point for which the leverage is being calculated \cite{Reference3}. Diagonal values in the H matrix represent the leverage values for different points in a given dataset. Data points far from the centroid will be associated with higher leverage.

\subsubsection{Confidence Estimation}
\paragraph{Standard Deviation}The standard deviation of the predictions, obtained from an ensemble of models, can be used as an estimator of model uncertainty for a given input \cite{Reference1}. In our work, we constructed a homogenous ensemble using the bagging technique. This standard deviation $\hat{\sigma}$ was found to correlate with prediction accuracy. The underlying principle is that if different models yield significantly different predictions for a particular data point, then the prediction for this data point is more likely to be unreliable. Therefore we can employ the sample SD value as an AD measure, where high AD values for a prediction mean low confidence and low AD values mean high confidence.

The mathematical formula to represent an ensemble of models is as follows.
\begin{equation}
F(x) = h(f_1(x),f_2(x)....f_n(x))  
\end{equation}
where F(x) represents the ensemble prediction for input x. $f_n(x)$ represents the prediction of the individual model i. $n$ represents the total number of individual models in the ensemble. $h$ represents the combining function, which is typically the average of the individual model predictions given as.
\begin{equation}
F(f_1(x), f_2(x), ..., f_n(x)) = (1/n)\sum f_i(x) 
\end{equation}

Henceforth, the SD of the prediction  for an input x is given as:

\begin{equation}
d_{std}(x)= F(x) =  \sqrt{\frac{\sum_{i=1}^{N} (f_i(x)-\overline{F(x)})}{N-1}}
\end{equation}

\paragraph{CORELL}
The basic concept of Correll is that it measures the correlation between the training set predictions and the test set predictions \cite{Reference1}. Similar to Standard Deviations this model is applicable to an ensemble of models.  
Mathematically CORREL for an input x with an ensemble of N models is defined as:
\begin{equation}
d_{correll}(x)= 1 - max_{\footnotesize{i=1,..,N}}[corr(\vec{f(T_i)},\vec{f(x)})]
\end{equation}
$\vec{f(T_i)}$ is the vector of training set prediction and $\vec{f(x)}$ is the test point prediction. $Corr$ is the Spearman correlation between the two vectors. The low value of CORREL (i.e., high Spearman correlation coefficient) indicates that for a test point x, there is a data point  $T_k$ in the training set for which predictions of the ensemble of models are strongly correlated. In other words, CORRELL checks if there is a correlation between the test data point and any data point in the training set. A high correlation means that the test data point is very similar to the training data and is within the applicability domain.

\paragraph{Gaussian Process Regressor} is a probabilistic model and implements Gaussian processes (GP) for regression purposes. It is a non-parametric approach that models the relationship between input variables and their corresponding output values. GPR assumes that the underlying function generating the data follows a Gaussian process, which is a collection of random variables that can be described by their mean and covariance. The prior mean is assumed to be constant and zero. The prior’s covariance is specified by passing a kernel object. The hyperparameters of the kernel are optimized during the fitting of the Gaussian ProcessRegressor by maximizing the log-marginal-likelihood (LML) based on the passed optimizer. As the LML may have multiple local optima, the optimizer can be started repeatedly. The method requires adjustment of three hyperparameters—alpha, which stands for the noise level (also acts as a regularization of the model), the parameter gamma of the RBF kernel which represents the covariance function, and variance threshold $\sigma^*$. The sckitlearn library was used for the GPR implementation \cite{Gprsklearn}.

The predictions of GPR are represented by a Gaussian distribution, characterized by a mean and a covariance. The mean represents the estimated value of the target variable, while the covariance captures the uncertainty or variability associated with that prediction. This variance can be used as a measure of AD measure \cite{Assima2020} just like the Standard Deviation method, where high variance means low confidence in prediction and low value of variance means high confidence in the prediction.

\paragraph{Random Forests} 
Random Forests, an ensemble learning method, combines multiple decision trees for predictions. Each tree is trained on a random subset of the data and features, promoting diversity and reducing overfitting. Predictions are made independently by each tree, and the final prediction is determined by aggregating all tree predictions. Leveraging the ensemble nature, we use Random Forest can be used as an applicability domain (AD) measure. By aggregating tree predictions and calculating the standard deviation across the ensemble, a higher standard deviation indicates a higher AD value, signifying low confidence in the prediction

\paragraph{Bayesian Neural Networks} This is a novel approach we introduced in the field of defining the applicability domain. Unlike traditional neural networks where weights are assigned as a single value or point estimate, in BNN the weights are considered a probability distribution. A fundamental characteristic of the Bayesian approach is its emphasis on marginalization rather than optimization \cite{Gordon2020Wilson}. In Bayesian methods, we consider and incorporate solutions from all possible parameter settings, taking into account their respective posterior probabilities. Instead of relying solely on a single parameter setting, the Bayesian approach assigns weights to each solution based on its probability, allowing for a more comprehensive representation of possible outcomes. By considering the full range of parameter settings and their associated probabilities, the Bayesian approach provides a more nuanced and probabilistic understanding of the problem, avoiding overreliance on a single parameter setting.

A Bayesian approach defines a full probability distribution over parameters known as the posterior distribution. The posterior represents our belief/hypothesis/uncertainty about the value of each parameter. We use the Bayes Theorem to compute the posterior.
\begin{equation}
    P(\theta|X) = \frac{P(X|\theta)P(\theta)}{P(X)}\label{eq:bayestheorm}
\end{equation}
Where $X$ is the data, $P(X|\theta)$ is the likelihood of observing $X$, given weights $\theta$, $P(\theta)$ is the prior belief of the weights, and the denominator $P(X)$ is the probability of observing the data over all the possible values of the parameters and it requires integrating over all possible values of the weights as 
\begin{equation}
P(X) = \int P(X|\theta)P(\theta)d\theta\
\end{equation}
"Integrating over the indefinite weights in evidence makes it hard to find a closed-form analytical solution. Therefore, simulation and numerical-based methods like Monte Carlo Markov Chain (MCMC) and variational inference (VI) are commonly employed. MCMC is a foundational approach in Bayesian statistics but can be slow for large datasets or complex models. In contrast, VI provides a faster alternative. VI identifies the closest probability distribution to the posterior, making it simple to work with. An optimization algorithm is then used to learn the parameters of this distribution, minimizing the divergence from the true posterior distribution. Formally, a new distribution $Q(\theta|z)$ approximates the true posterior $P(\theta|X)$, with VI optimizing parameters to minimize distribution divergence.
\begin{equation}
    Q^*(\theta) = argmin_zKL(Q \; || \; P)
\end{equation}

In the above equation, KL or KL Divergence is a non-symmetric measure of similarity or relative entropy between the two distributions \cite{Kullback1951Leibler}. We solve for the value of z so that it minimizes the KL between $P(\theta|X)$. $Q(\theta|z)$.The KL divergence between $P(\theta|X)$. $Q(\theta|z)$ is given as:

\begin{equation}
     KL(Q \; || \; P) = \int Q(\theta|z)Log\frac{Q(\theta|z)}{P(\theta|X)}d\theta\
\end{equation}

Replacing the $P(\theta|X)$ using equation \ref{eq:bayestheorm}, we get:\\
\begin{equation}
\begin{aligned}
KL[Q(\theta|z) \| P(\theta|X)] &= \int Q(\theta|z)\log \frac{Q(\theta|z)P(X)}{P(X|\theta)P(\theta)} \, d\theta \\
&= \int Q(\theta|z)\left[\log Q(\theta|z)P(X) - \log P(X|\theta)P(\theta)\right] \, d\theta \\
&= \int Q(\theta|z)\log \frac{Q(\theta|z)}{P(X)} \, d\theta + \int Q(\theta|z)\log P(\theta) \, d\theta \\
&\quad - \int Q(\theta|z)P(X|\theta) \, d\theta
\end{aligned}
\end{equation}

Taking the expectation to $Q(\theta|z)$, we get:
\begin{equation}
    KL[Q(\theta|z) \| P(\theta|X)] = \mathbb{E}\left[Log\frac{Q(\theta|z)}{P(\theta)}\right] + LogP(X) + \mathbb{E}[LogP(X|\theta)]
\end{equation}

The above equation still has the term $Log P(X)$, making it difficult for KL to compute. Therefore, an alternative objective function is derived by adding $LogP(X)$ with negative KL divergence. $LogP(X)$ is a constant with respect to $Q(\theta|z)$. The new function is called the evidence of lower bound (ELBO) which we optimize and is expressed as:

\begin{equation}
\begin{aligned}
ELBO(Q) &= \mathbb{E}[\log P(X|\theta)] - \mathbb{E}\left[\log \frac{Q(\theta|z)}{P(\theta)}\right]\\
&= \mathbb{E}[\log P(X|\theta)] - \text{KL}\left[Q(\theta|z) \| P(\theta)\right]
\end{aligned}
\end{equation}

The ELBO for a Bayesian Neural Network (BNN) is a trade-off between two terms. The first term, $\mathbb{E}[\log P(X|\theta)]$, known as the likelihood, ensures a good fit to the observed data. It measures the expected log-likelihood of the data given the weights. The second term, $\text{KL}\left[Q(\theta|z) | P(\theta)\right]$, penalizes the divergence between the variational and prior distributions over weights. This acts as a regularization term, preventing overfitting and promoting alignment with prior beliefs. Optimizing the ELBO through algorithms like gradient descent minimizes the KL divergence.

In our study, we implemented a 4-layer neural network architecture with probabilistic weights and biases. The model was trained for 200 epochs using the evidence lower bound (ELBO). We evaluated model performance on a test set using Mean Squared Error (MSE) and Kullback-Leibler (KL) divergence metrics. MSE gauged average squared differences between predicted and actual values, while KL divergence measured dissimilarity between predicted and true probability distributions. We utilized the \textit{torchbnn} library \cite{lee2022graddiv} for BNN development. To assess prediction uncertainty, we performed 1000 iterations for each test example, calculating standard deviation as an AD measure. Higher AD values indicated increased uncertainty and lower prediction reliability. Employing BNNs for both regression and AD methods enhanced overall efficiency by ensuring consistent learning of the underlying function. This methodology, iteratively validated, provided a comprehensive understanding of predictions and associated uncertainties.

\begin{figure}[th]
\centering
\includegraphics[width=8cm]{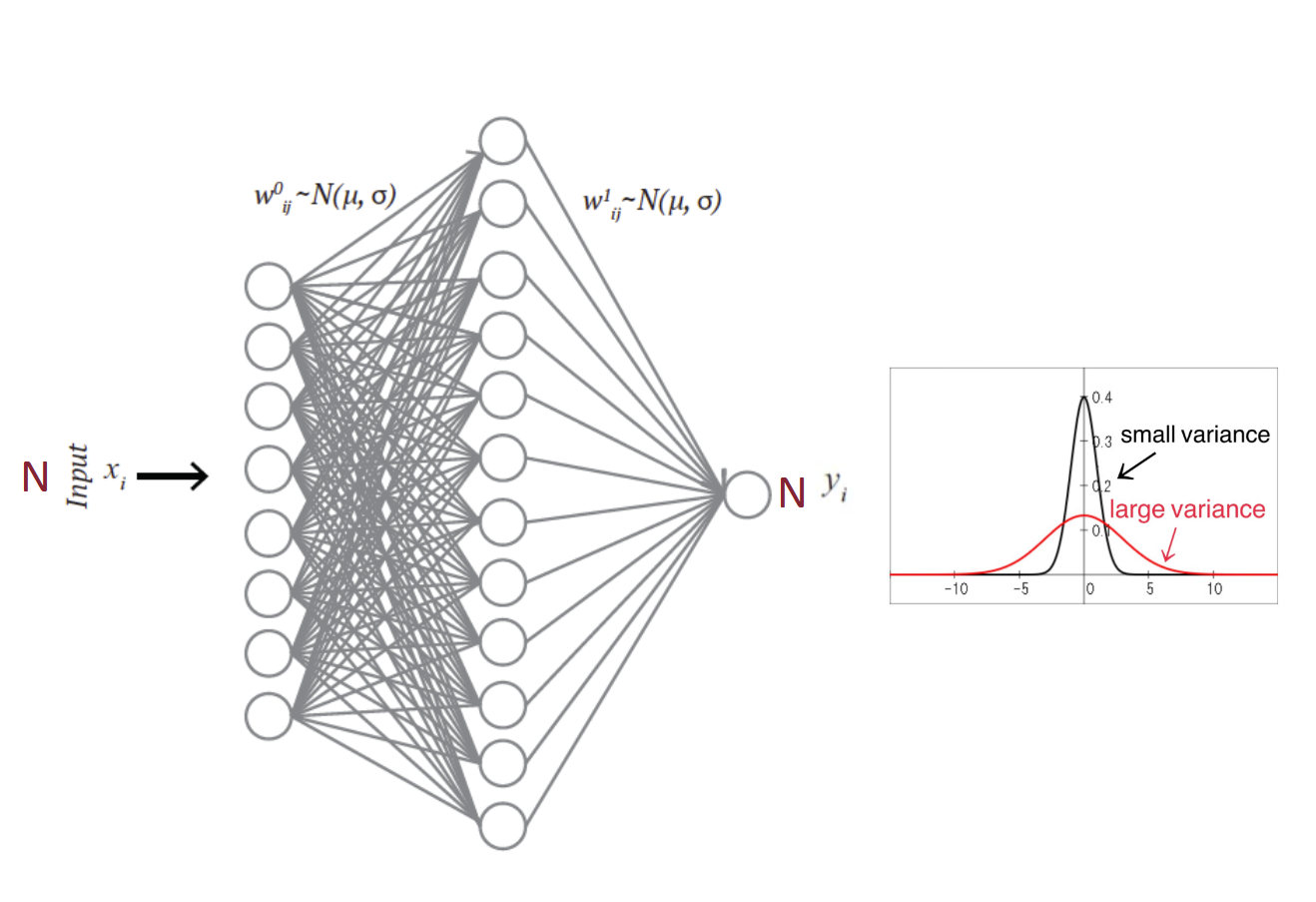}
\caption[Working of BNN]{Evaluating BNNs on a test set for n-iterations. A high variance in output indicates low confidence, while a low variance implies higher confidence. This figure is inspired by the works of \cite{Sabber2019}.}
\label{fig:BNN}
\end{figure}

\section{Method}
\begin{figure}[th]
\centering
\includegraphics[width=8cm]{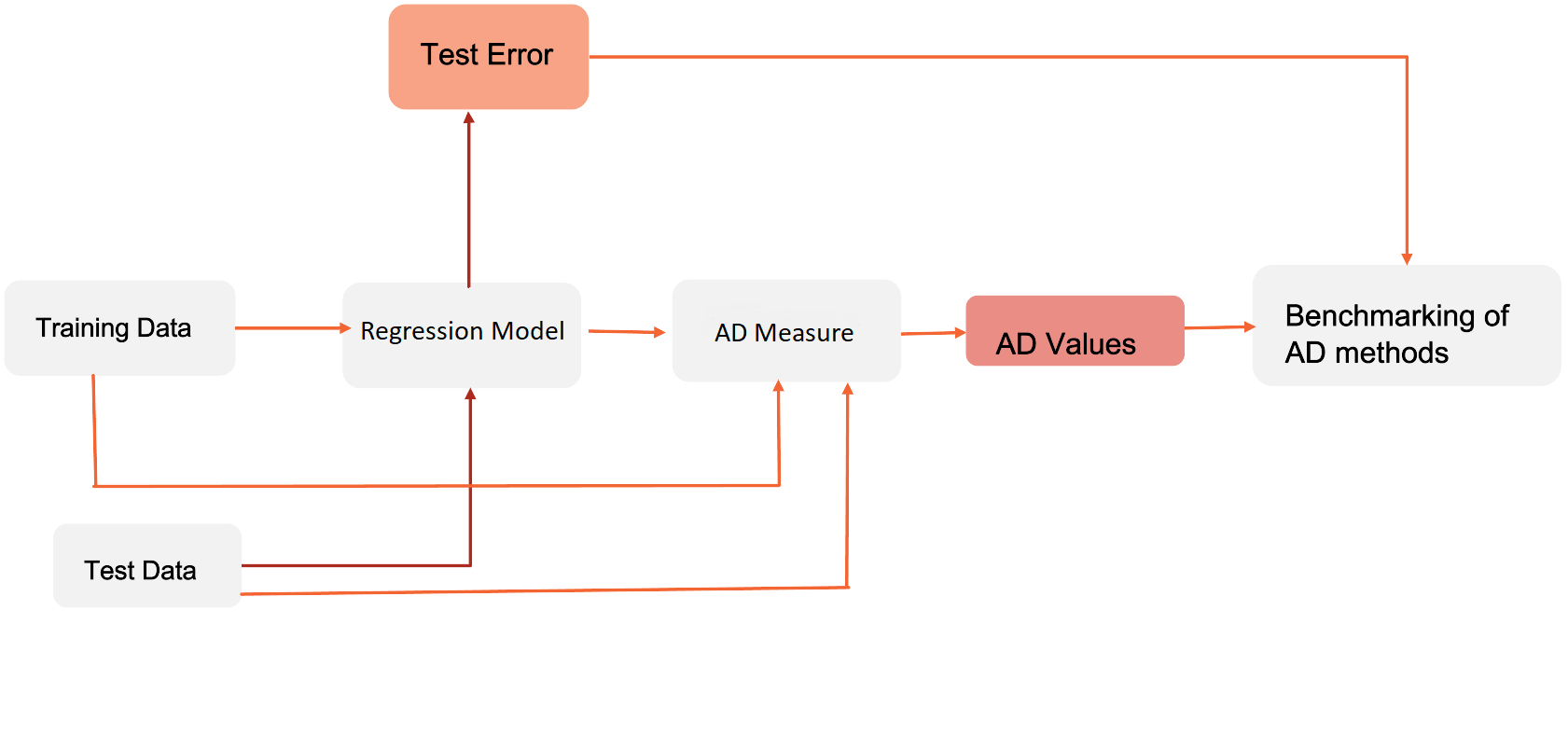}
\caption[Workflow]{The figure depicts our study's workflow. We start by training the regression model and evaluating it on the test set. Then we apply the AD measure to get the AD Values. The AD values along with the test error are used to assess the performance of the AD measure. We expected to observe a monotonic increasing trend between the AD values and the test error}
\label{fig:workflow}
\end{figure}

For our study, We focused solely on regression tasks and employed seven regression models as our prediction models, finally, we utilized a Bayesian neural network (BNN) as both the primary predictive model and the applicability domain (AD) measure. The regression models were trained on five publicly available datasets and evaluated on the corresponding test sets. We trained the regression models until convergence. Each model was evaluated on the test set using metrics such as root mean square error (RMSE) and R2 score. Additionally, we calculated the absolute error for each test point. After training the regression model different AD measures were applied to the regression models to define the applicability domain of the models. These AD measures were evaluated on the test set, providing AD values for each test point. Finally to benchmark the different AD methods we plot the absolute errors of the test points with their corresponding AD values and then use the validation framework to compare the performances of the AD methods. As mentioned before a good AD measure will be positively correlated with prediction errors, predictions with high absolute
error should have a high AD value and vice versa.
The overview of the workflow is shown in Fig \ref{fig:workflow}. In the diagram, we can see the regression model is trained on the training set until convergence. Then we apply
each of the AD measures. Finally, we evaluate the AD measures with respect to the absolute errors generated by the test set

\subsection{Dataset}
The experiments used publicly available datasets from the UCI machine learning repository and Kaggle to estimate prediction accuracy. The datasets included Energy Efficiency, Boston Housing, California Housing Prices, Abalone, and Red Wine Quality. These datasets provide information on building energy efficiency, housing prices, abalone shell physical attributes, and chemical properties of red wines.

The datasets were preprocessed to ensure quality and compatibility. This included data cleaning to identify missing values or outliers, feature selection to focus on relevant variables, normalization to ensure the data was consistent across variables and encoding categorical values. These steps were performed using techniques such as imputation, correlation analysis, z-score normalization, and binary encoding.
To evaluate the regression models and the AD measures, we split the datasets into training and test sets using a 7:3 ratio. This split was consistently applied across all the different regression models. The training set was used to train the models, while the test set was used to assess their performance and generalizability.

\subsection{Regression models} In our study, we employed seven regression models: Linear Regression, Lasso Regression, Ridge Regression, Decision Trees Regressor, XGBoost, MLP regressor, Support Vector Regression, and Random Forest Regressor. The majority of these models were implemented using default parameters from the sklearn library \cite{scikit-learn}. Our training process involved meticulous adjustment of model parameters to minimize test errors, with a focus on achieving optimal performance.

To assess the accuracy and goodness of fit for each regression model, we employed common metrics such as root mean square error (RMSE) and R2 score. These metrics provided valuable insights into the predictive capabilities of our models. Thorough analyses were conducted to fine-tune the training parameters, ensuring the models' reliability and robustness in addressing regression tasks. This rigorous approach aimed to obtain precise and dependable results in evaluating the performance of each regression model.

\subsection{Calculating Applicability Domain Measures}
During the evaluation of the regression model, we also calculate the absolute error for each data point. Typically, the error is averaged over the whole available set and reported as a single value for example RMSE and MAE. However, this approach does not reflect the complete information about the prediction accuracy. For a subset of data points, the prediction accuracy may be significantly higher than the average, while for some data points the model may completely fail to predict the target. Thus to get the overall picture of the model's accuracy we decided to evaluate the absolute error $|y_{true}- y_{pred}|$ for each data point in the test set. Sushko \cite{Reference1} referred to it as \emph{Variable accuracy}. Then we evaluate the AD measures on the test set. This will yield the AD values for each data point in the test set. We sort these AD values in ascending order and plot them with respect to the corresponding prediction error of the data point in the test set. Since a higher AD value implies the inability to predict the data point
correctly, thus for data points with high AD values, we expect the model’s errors for
that data point to also be high, giving us a monotonically increasing plot 


\subsection{Validation Framework - Benchmarking Criteria}

To compare different applicability domain (AD) measures, assessing their ability to distinguish between low and high-accuracy predictions is crucial. Each AD measure exhibits varying capabilities in discriminating predictions of high and low accuracy. 
To quantify and compare their performance, a validation framework is employed. This framework is based on the expectation that prediction errors of the regression model should increase monotonically with respect to the AD values. Specific metrics within this framework evaluate the performance and effectiveness of these methods. The concept of error and prediction accuracy will be used interchangeably, as they are related, with higher errors resulting in lower prediction accuracy and vice versa.

\subsubsection{Cumulative Error and Coverage Plot}
Cumulative averaging is a method of calculating the average of a sequence of values by summing up all previous values and dividing by the total number of values encountered, providing an overall trend or average value over time, smoothing out fluctuations and incorporating the entire history of values.

\begin{equation}
\label{eq:cumu_1}
\text{Cumulative Average}_n = \frac{1}{n} \sum_{i=1}^{n} x_i
\end{equation}

Where $n$ represents the cumulative average up to the $n$th value, $x_i$ represents the $i$th value in the sequence, and $n$ represents the total number of values encountered so far.\\

\begin{figure}[th]
\centering
\includegraphics[width=9cm]{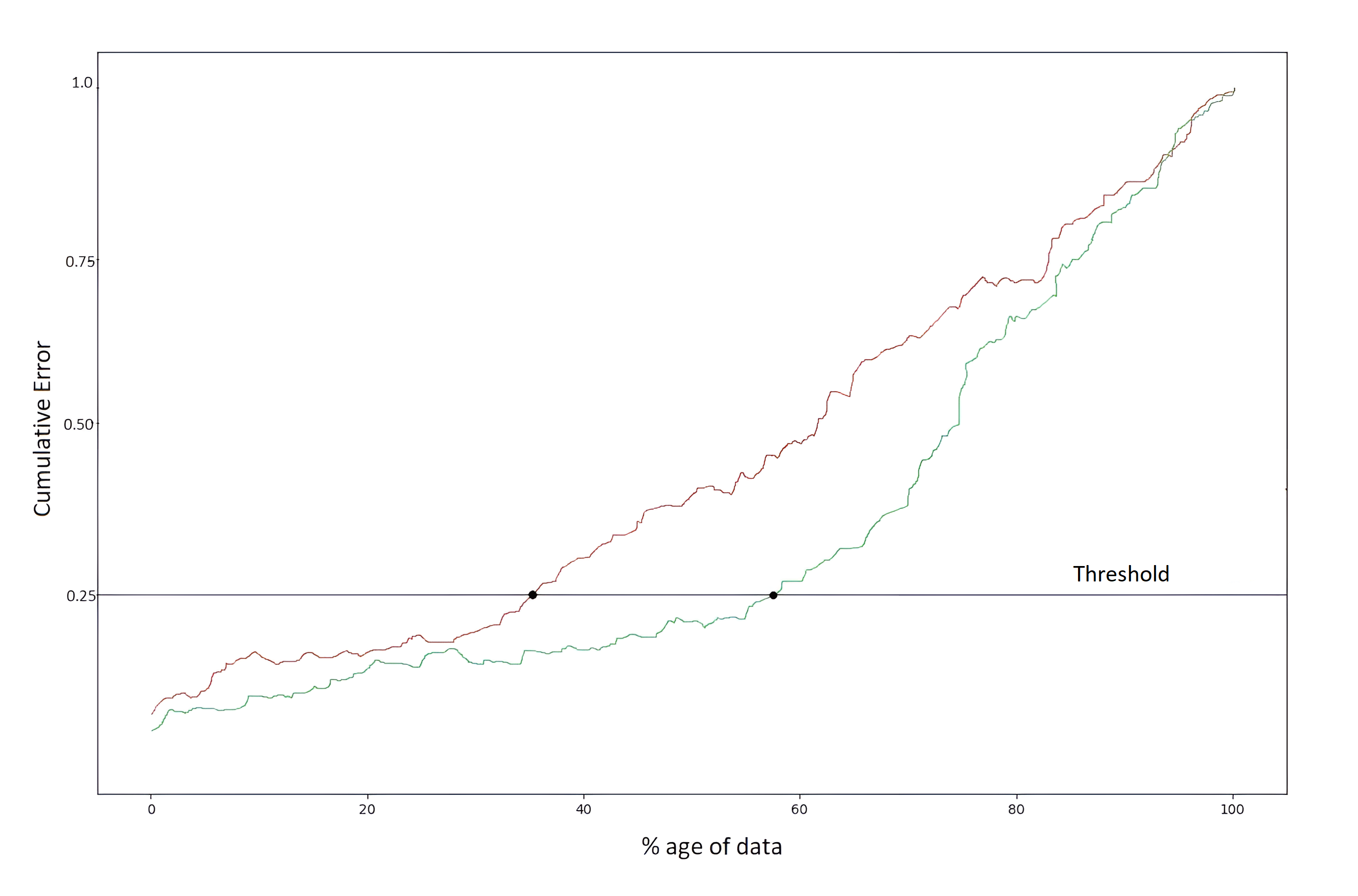}
\caption[Coverage Plot - Cumulative Error ]{An example of the coverage plot is illustrated, where the y-axis represents the cumulative error of predictions, and the x-axis depicts AD values on a percentage scale. We set the threshold at the 25th percentile of cumulative test errors. The coverage of the AD measure is measured by the percentage of data falling within this threshold. Notice the "red" AD measure outperforms the "green" since it covers roughly 60\% of the test set's data, compared to 35\% for the green.  }
\label{fig:cumulative}
\end{figure}

We sort the test set data points based on AD values, representing them on a \emph{percentage scale} instead of absolute values. The percentage scale indicates the proportion or percentage of test set data points with AD values at or below a given value. For instance, an AD value of "10\%" means that 10\% of the test set data have AD values equal to or less than this specific AD value. 

We then calculate the cumulative average of the prediction error of the test points using the equation. 
\begin{equation}
\label{eq:cumu_2}
   \frac{1}{n} \sum_{i=1}^{n} | y_{i_{\text{true}}} - y_{i_{\text{pred}}}|
\end{equation}
where n is the size of the test set.

The cumulative averaging in combination with the AD percentage scale results in a cumulative error or coverage plot \cite{Reference1}. 

This method employs a predetermined \emph{threshold} to decide if a data point is within the applicability domain of the model. The threshold is set at a specific percentile of absolute errors from test points for a given regression model.

Let n be the number of data points in the test set. The absolute error for each data point $i$ can be denoted as $\epsilon(i)$, defined as:

\begin{equation}
    \epsilon(i) = | y_{i_{\text{true}}} - y_{i_{\text{pred}}}|
\end{equation}

where
$ y_{i_{\text{true}}}$ represents the true value for data point i.
$y_{i_{\text{pred}}}$ represents the predicted value for data point i.
$\epsilon(i)$ represents the absolute error between the true and predicted values for data point i. Then we can find the cumulative error using the equation \ref{eq:cumu_1}. For instance, if we consider the 25th percentile of the cumulative errors in the total test set, we can define the AD  as the subset of data points whose cumulative errors are below this 25th percentile threshold.


The 25th percentile was consistently used as the threshold for AD evaluation in our experiments. We heuristically selected the 25th percentile as it provided an effective balance for our specific context, although it is not a standard threshold. This decision was based on the understanding that lower prediction errors correspond to higher accuracy. Setting the threshold too low could result in an overly inclusive AD, while setting it too high might exclude valuable data points with reasonably accurate predictions. The 25th percentile strikes this balance, allowing us to focus on data points with relatively lower prediction errors while maintaining a necessary level of strictness. Typically, the 25th percentile of total errors corresponds to approximately 75\% accuracy. To assess the performance of multiple AD measures, we averaged the obtained thresholds from these measures for each experiment. To evaluate AD measures using accuracy coverage, we followed these steps:

1. Sort the test set based on AD values, converting absolute values to a percentage scale.

2. Compute the cumulative average of the absolute errors for the data points in the sorted test set

3. Define a threshold and identify the AD value where the cumulative error is less than or equal to the threshold.

4. All the data points with AD values less or equal to the threshold are marked
inside the applicability domain. The percentage of test data that is included inside the threshold is called the \textit{Coverage}. Thus, for a given model a larger coverage value corresponds to an AD measure with larger numbers of reliable predictions.  (see Fig. \ref{fig:cumulative})





For each AD measure, we calculated the percentage of data points with an AD value less than the respective 25th percentile threshold for the test set, denoted as $C_{25}$ (C for coverage). $C_{25}$ serves as an estimate for the AD measure's performance, and measures are ranked based on their $C_{25}$ values. A higher $C_{25}$ value signifies a higher number of reliable predictions.

This cumulative averaging is easily interpretable and very stable against noise. However, it has two drawbacks. Firstly, it relies on the chosen accuracy threshold, potentially leading to different rankings for different thresholds. Secondly, accuracy coverage is influenced not just by an AD measure's ability to distinguish highly accurate predictions but also by the overall performance of the analyzed model. Models with higher prediction accuracies are likely to exhibit higher accuracy coverages for a similar threshold.

\subsubsection{Moving Average - Area Under the Curve}
Another metric \cite{Reference1} in our study is the area under the curve (AUC), calculated as the absolute difference between the moving average curve and the average model performance line. 
In the moving average plot, AD values are plotted against corresponding absolute errors for each data point in the test set. As per the nature of AD values, accuracy should not increase as AD values increase. 

This method calculates a symmetric moving average for a 1D array (sorted AD values, in this case) using a centered window of a specified odd size. It determines the half-window size for symmetric data consideration, assigns uniform weights to all points within this window, and employs reflective padding to manage edge effects effectively. The data is then convolved with these weights to produce a smoothed version of the original array, effectively reducing noise and smoothing out fluctuations in the data series.

\begin{figure}[th]
\centering
\includegraphics[width=10cm]{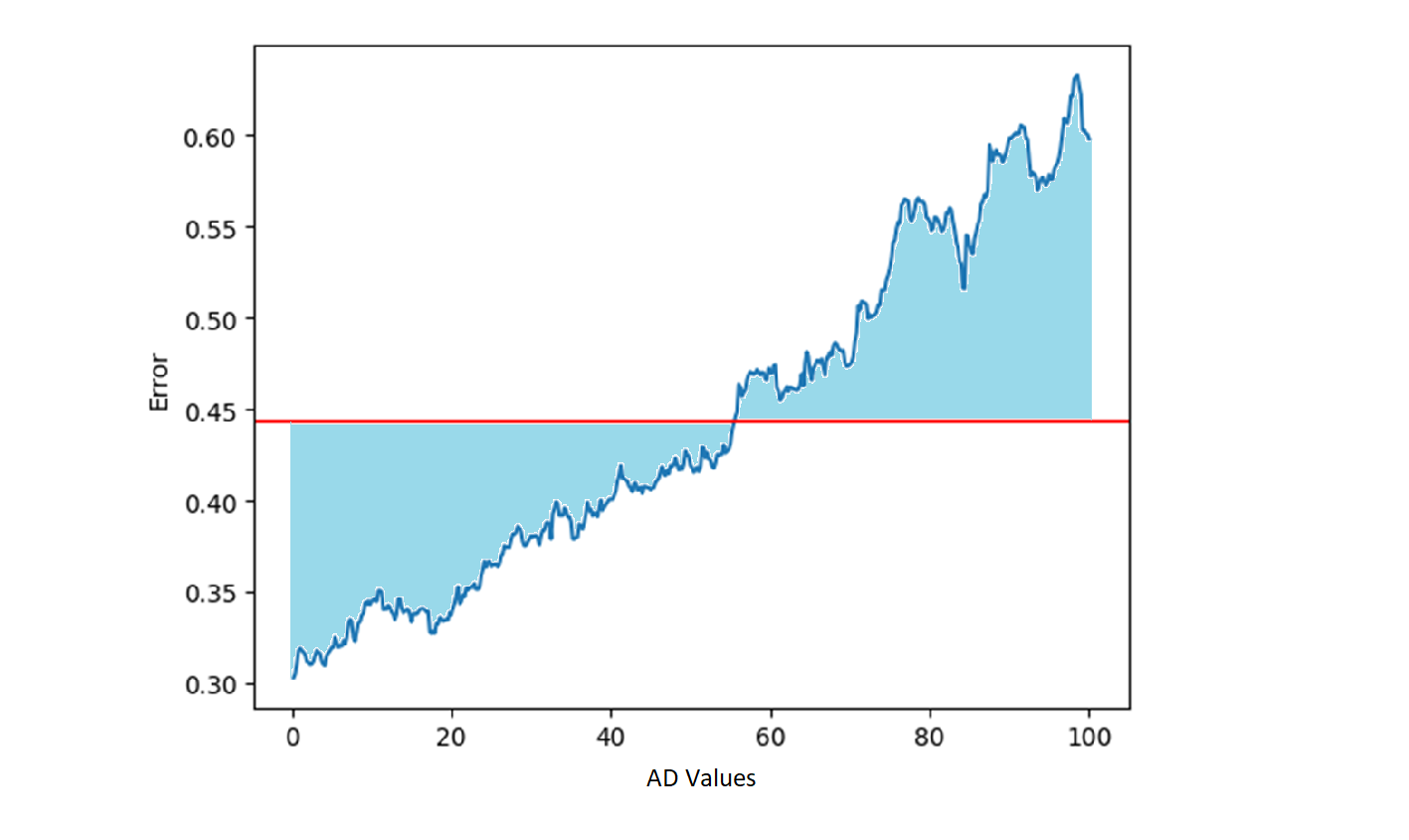}

\caption[AUC]{The area-under-curve (AUC) criterion corresponds to the filled area between the moving average plot and the average error of the model (the red horizontal line).}
\label{fig:auc}
\end{figure}

An example is shown in \ref{fig:auc}. We smooth the plot with a window size of N to reduce noise. Different window sizes were used for different datasets, depending on the size of the test set and treated as a hyperparameter. The window size must be chosen such that it reduces excessive noise while preserving critical information. We used visual analysis to determine the appropriate window size for each dataset. AD values were converted to percentage scales for standardized comparison across datasets and enhanced interpretability. 

The advantage of AUC is that unlike coverage we do not need to manually set up a threshold value. In Fig \ref{fig:auc}, this corresponds to the area between one of the solid lines and the red horizontal line. Formally,

\begin{equation}
\sum_{i=1}^{n} |\epsilon(i) - E_{avg}|
\end{equation}

where i is the AD value in the absolute scale and $\epsilon(i)$ is the absolute error for this AD value, $ E_{avg}$ is the global average error of the model, n is the no of examples in the test set. In our evaluation, we utilized the Mean Absolute Error (MAE) as the metric for calculating the average error.

This quantifies the total deviation of individual data points from a specified average value,  $ E_{avg}$. Although this metric provides a cumulative measure of deviation akin to the AUC, it is not a true AUC calculation since it does not account for the spacing between the data points, a critical factor in true area computations in graphical analysis. Nevertheless, it serves a similar purpose by offering an aggregate indication of variation from the expected norm.


\begin{table*}[t]
\centering
\caption{Average Ranks of the AD measures ranked by the percentages of test data covered for a threshold of the 25th percentile of the error. The coverage values for an AD measure are averaged across the dataset.}
\label{tab:coverage_ranks}
\begin{tabular}{lllllll} 
\toprule
AD Measure & Calif. & Boston & Energy & Abalone & Wine & \textbf{Avg. Covg.}\\
\midrule
\textbf{standard deviation} & 47.76 & 68.20 & 74.78 & 66.88 & 52.92 & \textbf{62.11} \\ 
leverages & 67.58 & 60.29 & 26.68 & 57.87 & 27.06 & 47.90 \\ 
kappa & 36.25 & 71.60 & 29.55 & 53.56 & 39.74 & 46.14\\ 
gamma & 33.33 & 69.13 & 33.39 & 50.08 & 44.51 & 46.09\\ 
\textbf{bnn} & 66.08 & 40.90 & 55.36 & 26.32 & 34.56 &\textbf{ 44.64 }\\
gpr & 9.26 & 69.56 & 10.01 & 64.01 & 28.05 & 36.18\\ 
random forest & 27.47 & 38.44 & 46.97 & 0.00 & 37.47 & 30.07\\ 
min\_kappa & 29.93 & 34.44 & 27.54 & 8.56 & 25.76 & 25.25\\
cosines & 5.17 & 10.03 & 62.34 & 5.26 & 31.67 & 22.89 \\
delta & 6.09 & 13.61 & 30.57 & 0.58 & 0.73 & 10.32\\ 
correll & 1.39 & 7.48 & 0.92 & 3.62 & 10.34 & 4.75 \\
\bottomrule
\end{tabular}
\end{table*}

 \section{Results}

\paragraph{Comparing different AD Measures }The evaluation of various applicability domain measures was conducted in conjunction with seven regression models to determine the most effective one. In the case of Bayesian Neural Networks, we initially trained it solely as an AD measure using it in conjunction with another regression model. Subsequently, we explored its dual functionality—using it as a regression model and using its built-in confidence estimates as AD measures. This dual role exhibited exceptional performance, yielding promising results in our experimental analysis.

This evaluation aimed to rank AD measures for each regression model across multiple datasets, determining the most effective measure based on performance metrics. The study also investigated whether confidence or novelty measures were more effective in distinguishing reliable from less reliable predictions.

The comparison of applicability domain (AD) measures was based on two criteria: \textit{accuracy coverage } and \textit{area under the curve (AUC)}. The evaluation followed a two-step approach.


First, we computed the average coverage and AUC values of an AD measure for each dataset, using all the regression models. Then this process was repeated for all datasets, giving us a final averaged score that reflected how well the AD measure performed on various models and datasets. This score was a comprehensive indicator of the effectiveness of the AD measure.




\paragraph{Comparison using coverage }Table \ref{tab:coverage_ranks} displays the final coverage value $C_{25}$ of all AD measures, averaged across all datasets. Each column represents a specific dataset and the cells within the column represent the averaged coverage value of an AD measure across all regression models it was applied to. The final coverage value is obtained by averaging the coverage values of the AD measure across all datasets and is presented in the last column: \textit{Avg. Covg}.

It can be seen in Table \ref{tab:coverage_ranks}, sd\_model achieved the highest performance, covering 63.14\% of test data across all regression models and datasets. Following closely were kappa and leverages with coverages of 45.9\% and 45.2\%, respectively. Correll exhibited the poorest overall performance. Bayesian NN had a 44.05\% coverage. Interestingly, BNN's performance as a standalone AD measure was not as expected, but subsequent sections reveal improved performance when used as both a regression model and an AD measure.


\begin{table*}[t]
\centering
\caption{AUC score for all the AD measures}
\label{tab:auc_scores}
\begin{tabular}{lllllll}
\toprule
AD Measure & Calif. & Boston & Energy & Abalone & Wine & \textbf{Avg. AUC} \\
\midrule
\textbf{standard deviation} & 0.71 & 1.00 & 1.00 & 1.00 & 1.00 & \textbf{0.94}\\
\textbf{bnn} & 1.00 & 0.61 & 0.76 & 0.85 & 0.62 & \textbf{0.77}\\
gpr & 0.22 & 0.94 & 0.29 & 0.69 & 0.54 & 0.54\\
gamma & 0.09 & 0.84 & 0.29 & 0.50 & 0.93 & 0.53\\
kappa & 0.09 & 0.92 & 0.28 & 0.54 & 0.66 & 0.50\\
leverages & 0.29 & 0.62 & 0.34 & 0.65 & 0.52 & 0.48\\
random forest & 0.59 & 0.42 & 0.61 & 0.18 & 0.60 & 0.48\\
min\_kappa & 0.06 & 0.51 & 0.00 & 0.38 & 0.88 & 0.37\\
cosines & 0.01 & 0.00 & 0.39 & 0.16 & 0.87 & 0.29\\
delta & 0.00 & 0.03 & 0.28 & 0.42 & 0.09 & 0.16\\
correll & 0.03 & 0.00 & 0.16 & 0.00 & 0.00 & 0.04\\
\bottomrule
\end{tabular}
\end{table*}

\paragraph{Comparison using AUC } Apart from the coverage we also incorporated AUC as the criterion to compare the performances. Table \ref{tab:auc_scores} summarizes the average AUC values. Since the range of errors was on a different scale for different regression models, we scaled the values of the area using \textit{min-max scaling} to get the AUC values on the same scale. In Table \ref{tab:auc_scores} it can be seen that the differences in AUC scores between the best and worst AD measure for each data set. We can see that in terms of AUC also, the sd\_model came on top with an average AUC of 0.45 this was closely followed by BNN with an AUC value of 0.4. CORRELL in this case also had the worst performance ranking last with an average AUC of 0.12. 

Within the group of novelty measures the best-performing AD measure is gamma closely followed by kappa and leverages. It is worth noting that these measures exhibit nearly identical scores, indicating their comparable performance levels.

\begin{figure*}[th]
  \centering
  \begin{subfigure}{0.45\textwidth}
    \centering
    \includegraphics[width=0.99\linewidth]{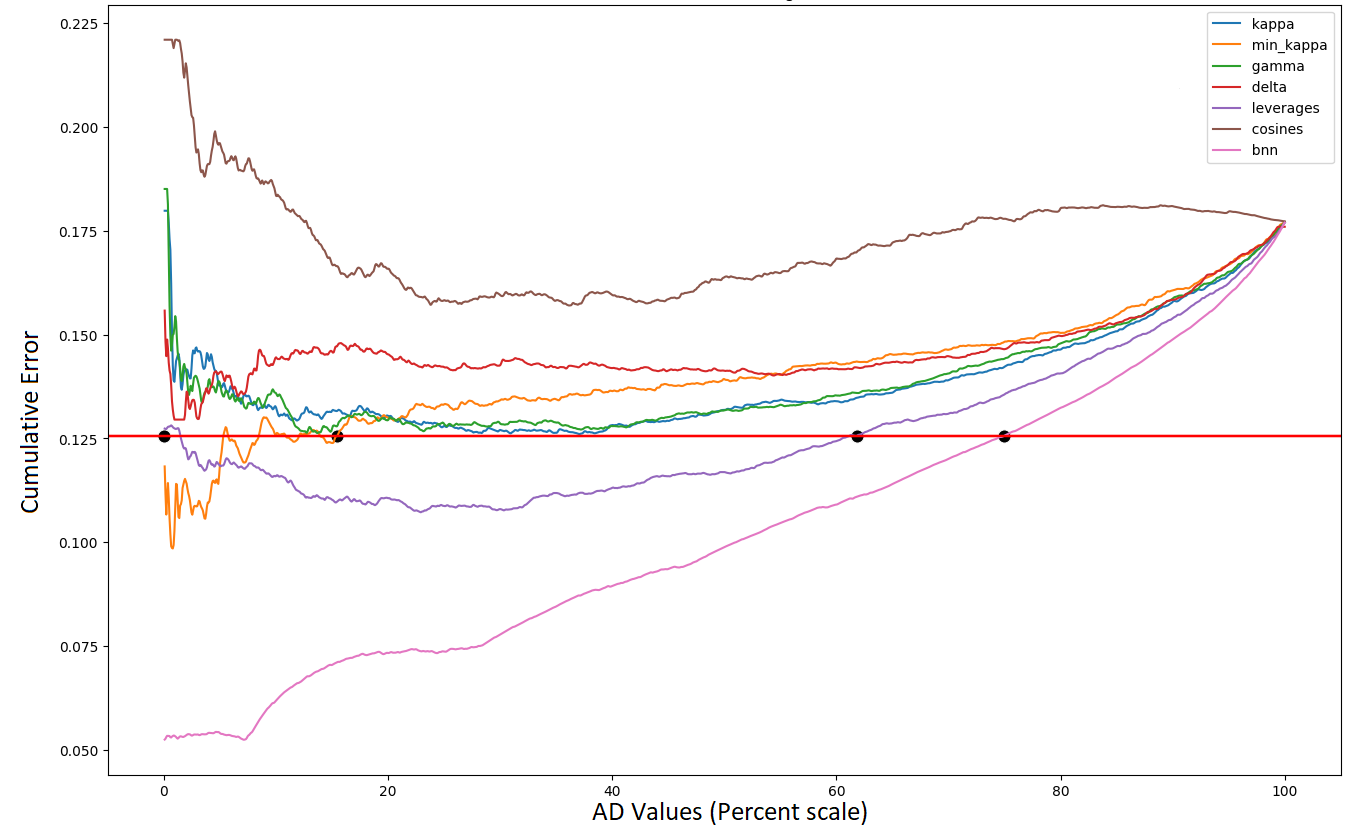}
    \caption{Coverage}
    \label{subfig:auc1}
  \end{subfigure}
  \hfill 
  \begin{subfigure}{0.45\textwidth}
    \centering
    \includegraphics[width=0.98\linewidth]{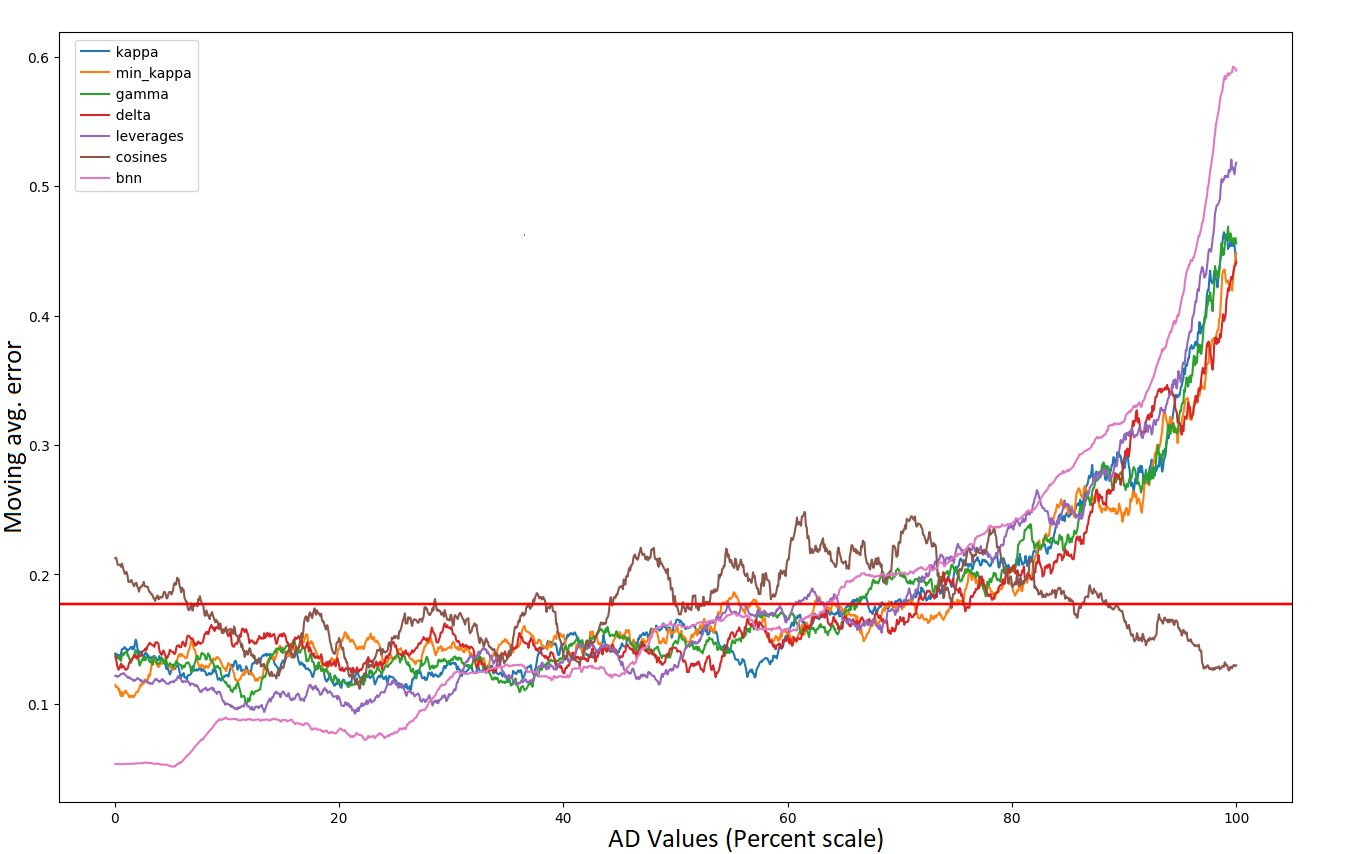}
    \caption{AUC}
    \label{subfig:auc2}
  \end{subfigure}
  \caption{The BNN, utilized as a regression model on the Abalone dataset with built-in confidence estimates as Applicability Domain (AD) measures, showcases superior performance compared to other novelty-based AD measures. BNN covers almost 80\% of the test data with reliable accuracy. Also, the moving average plot of the BNN exhibits a decent monotonically increasing curve with respect to the error values. This instance is one of several, as the model was applied the model to multiple datasets. The overall results are shown in Table \ref{tab:BNN_scores}.}
  \label{fig:auc2}
\end{figure*}

\paragraph{BNN's Dual Role: Regression Model and Applicability Domain Measure}
When BNN was used as a regression model and its built-in confidence estimates were used as an AD measure it demonstrated exceptional performance and promising results in determining the applicability domain. This is because, when the BNN is utilized as both a regression model and an applicability domain method, it leverages the same neural network architecture, weights, and connections for both tasks. Thus the model that is making the prediction is also measuring the AD, resulting in better performance. Table \ref{tab:BNN_scores} shows the coverage and AUC scores of BNN when implemented in this manner. The scores are averaged for all the datasets to give the final mean score. We can see the final coverage value of \emph{bnn} is 67.92\% which is the highest that all the previous AD measures benchmarked. Also, the average AUC value is 0.88 which is just second to the \emph{standard deviation}. These values show that our method exhibits greater performance compared to other AD measures. 
This highlights that Bayesian Neural Networks (BNN) demonstrate much better results when used with built-in confidence estimation. On the other hand, when BNN is employed as an AD measure with a different regression model, where a distinct model generates predictions, it leads to a performance drop. The mismatch between the prediction model and the AD measure causes poorer AD detection performance
\begin{sloppypar}
\begin{table}[t]
\centering
\caption{Coverage and AUC Scores for \textbf{BNN} when used as a regression model with its built-in confidence estimates as AD measure.}
\label{tab:BNN_scores}
\small 
\begin{tabular}{l*{6}{p{0.8cm}}} 
\toprule
\textbf{BNN} & \textbf{Calif.} & \textbf{Boston} & \textbf{Energy} & \textbf{Abalone} & \textbf{Wine} & \textbf{Avg.} \\
\midrule
\textbf{Coverage} & 81.07 & 38.7 & 79.65 & 74.98 & 65.2 & \textbf{67.92} \\
\textbf{AUC} & 1.00 & 0.41 & 1.00 & 1.00 & 1.00 & \textbf{0.88}\\
\bottomrule
\end{tabular}
\end{table}
\end{sloppypar}

Based on this observation, we can infer that models equipped with built-in confidence estimation (bnn) or an ensemble of homogeneous models (standard deviation), when utilized as Applicability Domain (AD) measures, tend to outperform other methods. There is a significant leap in the performance of these methods compared to the other methods, as demonstrated in Table \ref{tab:coverage_ranks}, \ref{tab:auc_scores}, \ref{tab:BNN_scores}

Based on the findings of our study, we recommend two approaches for defining the Applicability Domain (AD) in regression problems. Firstly, utilizing an ensemble of the same model used for prediction allows for improved AD estimation. This approach leverages the collective knowledge of multiple instances of the same model
to enhance reliability and capture a wider range of potential outliers or the uncertain-
ties. Secondly, employing models with built-in confidence estimates, such as those
based on Bayesian Frameworks, offers a robust AD definition. These models inherently provide measures of confidence or uncertainty, enabling more accurate identification of unreliable predictions and enhancing the reliability of the AD estimation
process. By adopting these recommended approaches, researchers and practitioners
can effectively define the AD and improve the overall performance and reliability of
regression models.


\section{Conflict of Interest}
This work was sponsored by GlaxoSmithKline Biologicals SA. Matthieu Duvinage is employee of the GSK
group of companies. Shakir Khurshid, department of Computer Science, Sapienza University of Rome. Bharath Kumar Loganathan is an employee of Cognizant Technology Solutions, a technical consultancy
firm contracted by GlaxoSmithKline Biologicals SA in the context of this work.
\section{Conclusions}
This study focuses on the Applicability Domain (AD) problem in machine learning models, crucial for understanding the limitations and validity of predictions based on input data characteristics. Two key approaches are suggested for defining AD in regression problems: novelty detection to flag unusual data points and confidence estimation to identify unreliable predictions. The evaluation of eight AD measures, including a novel Bayesian Neural Network (BNN) approach, revealed that the standard deviation of the ensemble of a regression model consistently performed best and BNN when used as both a regression model and an AD measure outperformed other methods in Coverage and was second to only \textit{standard deviation} in AUC scores. The study recommends employing ensembles of the same model for improved AD estimation and using models with built-in confidence estimates, like Bayesian Frameworks, for robust AD definition. These approaches enhance overall performance and reliability in regression models by effectively defining the AD. The variance in ensemble or BNN predictions serves as an indicator of generalization capability, highlighting areas of uncertainty and potential data points outside the model's applicability. Higher variance indicates the model lacks certainty for a specific data point, suggesting it's outside its applicability domain, While lower variance indicates higher confidence, potentially classifying the data point inside the model's applicability domain.

\vspace{12cm}
\normalsize
\bibliography{references}

\end{document}